\documentclass[runningheads]{llncs}

\usepackage[colorlinks,urlcolor=blue,linkcolor=blue,citecolor=blue]{hyperref}
\usepackage{color,array}
\usepackage{graphicx}

\usepackage{url}
\usepackage{hyperref}
\usepackage{graphicx}
\usepackage{subcaption}

\usepackage{booktabs} 
\usepackage{siunitx}  
\usepackage{tabularx} 

\def\citep{\cite}
\def\citet{\cite}

\begin{document}

\title{Left/Right brain, human motor control and the implications for robotics}

\author{Jarrad Rinaldo\inst{1} \and Jason Friedman\inst{2}\orcidID{0000-0001-8845-5082} \and Levin Kuhlmann\inst{1}\orcidID{0000-0002-5108-6348} \and Gideon Kowadlo\inst{1,3}\orcidID{0000-0001-6036-1180}}

\authorrunning{J. Rinaldo et al.}

\institute{Department of Data Science and AI, Monash University \email{levin@monash.edu.au} \and
Department of Physical Therapy, Tel Aviv University \email{jason@tau.ac.il} \and
Cerenaut \email{gideon@cerenaut.ai}\\
\url{https://cerenaut.ai}}

\maketitle


\begin{abstract}
    Neural Network movement controllers promise a variety of advantages over conventional control methods, however, they are not widely adopted due to their inability to produce reliably precise movements. This research explores a bilateral neural network architecture as a control system for motor tasks. We aimed to achieve hemispheric specialisation similar to what is observed in humans across different tasks; the dominant system (usually the right hand, left hemisphere) excels at tasks involving coordination and efficiency of movement, and the non-dominant system performs better at tasks requiring positional stability. Specialisation was achieved by training the hemispheres with different loss functions tailored to the expected behaviour of the respective hemispheres. We compared bilateral models with and without specialised hemispheres, with and without inter-hemispheric connectivity (representing the biological Corpus Callosum), and unilateral models with and without specialisation. The models were trained and tested on two tasks common in the human motor control literature: the random reach task, suited to the dominant system, a model with better coordination, and the hold position task, suited to the non-dominant system, a model with more stable movement. 
    Each system outperformed the non-preferred system in its preferred task. 
    For both tasks, a bilateral model outperformed the non-preferred hand and was as good or better than the preferred hand. 
The results suggest that the hemispheres could collaborate on tasks or work independently to their strengths.
This study provides ideas for how a biologically inspired bilateral architecture could be exploited for industrial motor control.
\end{abstract}

\keywords{Bi-hemispheric \and Bilateral \and Hemispheric specialisation \and Hemispheric asymmetry \and Deep Learning \and Motor control} 


\section{Introduction}
Controlling limbs is a crucial skill for humans and other animals. 
Part of this control system is the structure of our brains; two largely independent hemispheres with a Corpus Callosum sending signals between them. 
An advantage of the bilateral architecture is the ability for each hemisphere to specialise in different ways in different tasks \cite{allen_models_1983,goldberg2009new}. 

In human motor control, hemispheric specialisation presents as `handedness': the phenomenon seen in 96\% of the human population \cite{sainburg_handedness_2005}, causing a person to have a dominant and non-dominant side of the body. Handedness is a well-studied area that has explored why we have dominant sides of the body and what the differences are between the dominant and non-dominant sides.


Humans and other animals have exquisite control of their limbs. The bilateral architecture is conserved through millions of years of evolution, suggesting that it is an important principle that could be exploited for industrial motor control.
Conventional motor control in industry, without ANNs, often results in laborious setup processes involving the use of CAD, precise choreography of a task, or in some cases manual guiding of a robot \cite{hagele_industrial_2016}. ANN architectures offer promising improvements in flexibility \cite{shridhar_cliport_2021} but there are two main reasons for the lack of adoption. First, the most common uses for robotics in industrial settings are in manufacturing, where adaptability is less important as the environment is highly predictable; the setup process is arduous, but it rarely needs to be repeated. Second, ML/AI based models cannot be relied upon for pinpoint precision. 

Evidence suggests that motor control in humans does not suffer the same problems, due to the interaction and division of tasks between the dominant and non-dominant systems. \citet{sainburg_handedness_2005} concluded the dominant system (generally left brain, right arm) shows more coordinated movement, resulting in a more efficient trajectory, whereas the non-dominant system excels at positional stability of the limb. From these findings, Sainburg came to the conclusion that humans evolved this behaviour to be more capable in a wider range of tasks. Similar conclusions were reached by \citet{woytowicz_handedness_2018}, where in a cutting bread-like task it was shown that the dominant system was better at `cutting straight, while the nondominant system performed better at stabilisation or `holding the bread'. 


There are two aims. First, to better understand bilateral hemispheric specialisation in motor control; in particular, do specialised hemispheres improve performance and if so, how? 
We have 2 hypotheses: that they collaborate on given tasks and/or work independently for different tasks.
Second, we aim to use this understanding to build a more capable model than possible with standard techniques.

The objective of this research is to create a motor controller with a bilateral neural network that takes advantage of the specialisation seen in the human brain. 
Although the underlying cause of the differences in motor control properties between the dominant and non-dominant hemispheres is not well understood, in this study we used carefully selected loss functions to mimic these properties.
We test these models against non-specialised and unilateral equivalents to assess the usefulness of the approach. 

\subsection{Related work}
  
Hemispheric specialisation has been modelled with ANNs in a range of tasks such as word recognition \cite{weems_hemispheric_2004}, facial recognition \cite{dailey_organization_1999} and the line bisection task (where subjects are required to find the midpoint of a line) \cite{monaghan_cross-over_1998}. 
 No research has induced the hemispheric specialisation of motor control observed in animals and described by \citet{sainburg_handedness_2005} in an ML/AI model. This paper addresses this research gap.

Our approach could be seen as a special case of mixture of experts (MoE), motivated by biological analogy. However, there are some important differences. In a MoE system, the models typically do not interact, whereas in our bilateral system, the hemispheres do interact during training and inference, via the Corpus Callosum.

\section{Methodology}
\label{sec:method}

We used the framework provided in the MotorNet toolbox \cite{codol_motornet_2023} for the supervised training/testing harness, to simulate the robotic arm (referred to as the `plant') and to create novel tasks. Our source code is available at 
\url{https://github.com/Cerenaut/bilateral-motor-control}.

\subsection{Robotic manipulator}
\begin{figure*}
    \centering
    \begin{subfigure}[t]{0.3\textwidth}
        \centering
	    \includegraphics[width=\textwidth]{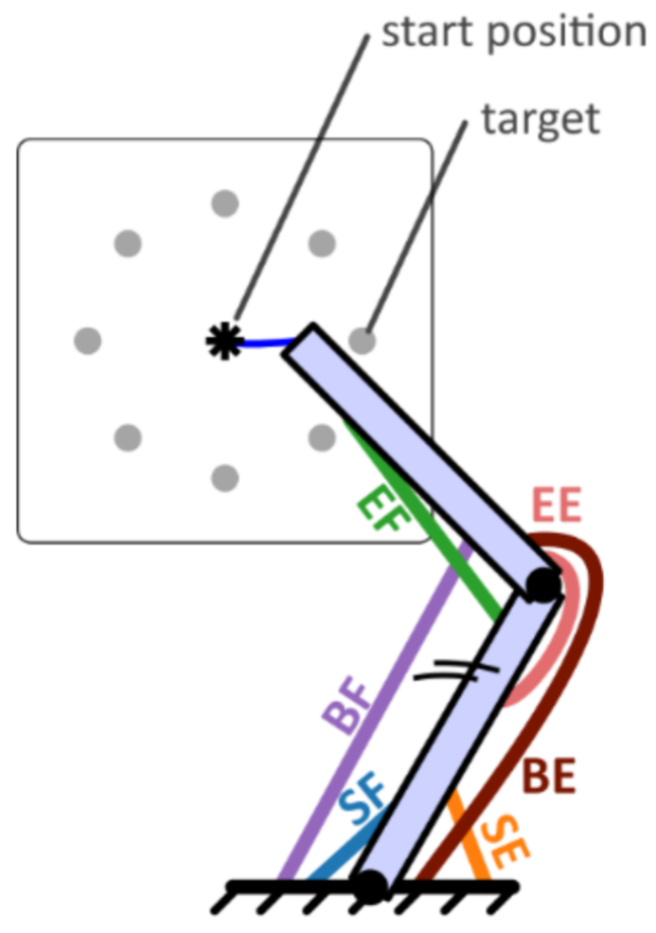}
    	\caption{}
	    \label{fig:plant}
    \end{subfigure}%
    ~ 
    \begin{subfigure}[t]{0.7\textwidth}
        \centering
	    \includegraphics[width=\textwidth]{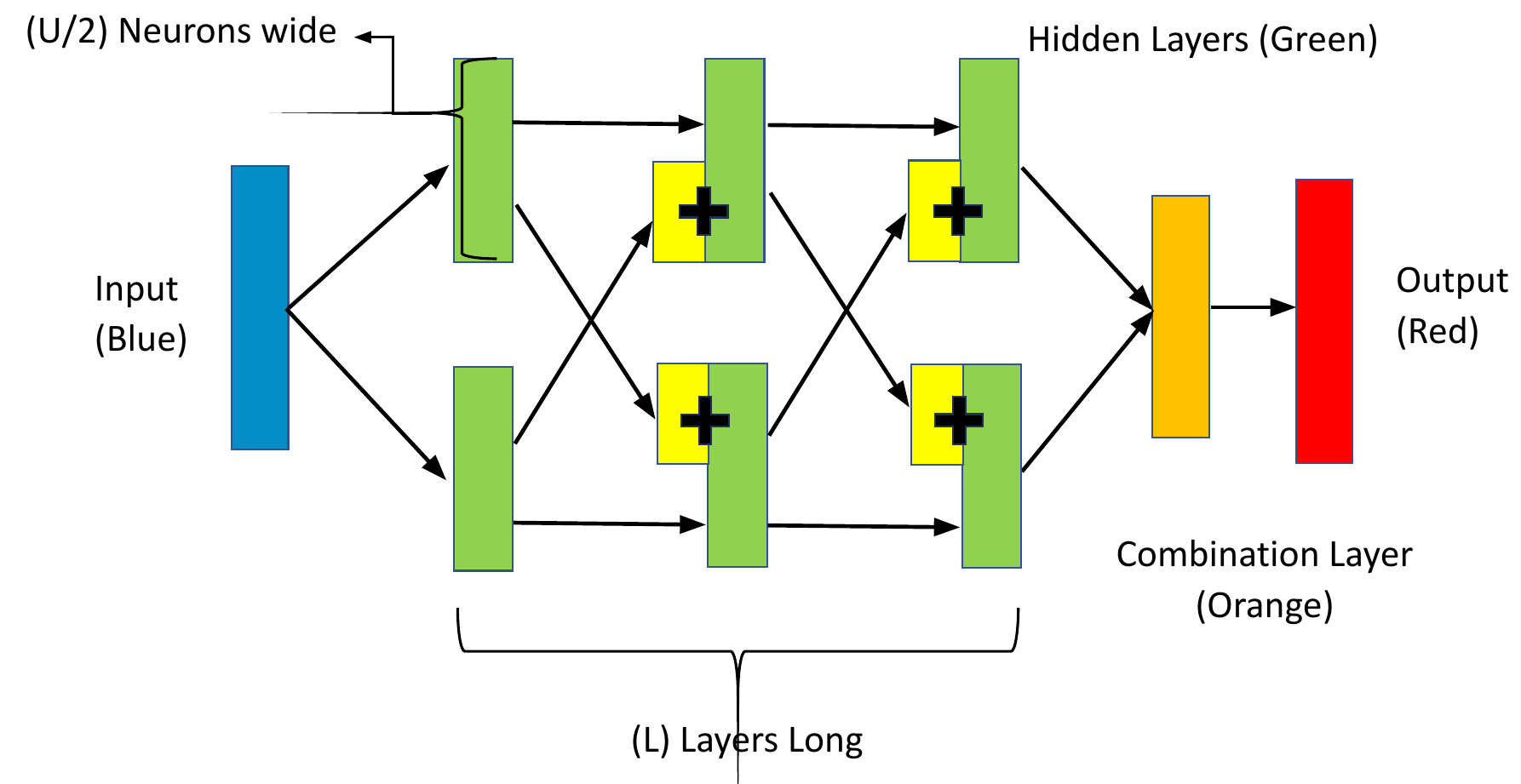}
    	\caption{}
	    \label{fig:arch_cc}
    \end{subfigure}
    \caption{a) MotorNet RigidTendon26 Plant \cite{codol_motornet_2023}. EE and EF refer to the elbow extensors and flexors, SE and SF to the shoulder extensors and flexors, BE and BF to the bi-articular extensors and flexors. b) Bilateral network with a Corpus Callosum.}
    \label{fig:arm_arch}
\end{figure*}

We used the RigidTendonArm26 model (with RigidTendonMuscle muscles) as it best represents a human arm, Figure~\ref{fig:arm_arch}a. The plant has 6 controllable muscles that are used to manipulate a 2-bone skeleton and provides two types of feedback: visual and proprioceptive. Visual feedback includes the current endpoint position that emulates what a human would see. Proprioceptive feedback gives information about the current state of its muscles, including the current length of the muscle and the current velocity of the muscle, similar to the human position sense. 


\subsection{Tasks}

We tested the model on two standard tasks in human motor control studies.
Task instances are dynamically generated using the MotorNet framework.
The tasks favour dominant and non-dominant systems, which allows us to observe how the models perform under different demands, where differences in hemispheric specialisation should be most apparent.

\begin{enumerate}
  \item \textbf{Random Reach Task:} The plant (arm) starts at a random position and is required to move the endpoint (hand) to a randomly generated target position, \textit{which favours the dominant hand}. No external forces were imposed, as this is more suited to the non-dominant hand. Accuracy is defined by the following.
  \begin{enumerate}
	\item \textbf{Goal Completion} Counts if the plant was able to successfully reach the target point (within a small threshold) and remain there for at least three timesteps.
    \item \textbf{Speed to Goal} measured as the distance from the start to the target location, divided by the timesteps until the goal is reached. 
  \end{enumerate}
  \item \textbf{Hold Position Task:} hold the position of the hand at a random target point (within a small threshold) in opposition to external forces with random direction and random bounded intensity, \textit{which favours the non-dominant hand}. Accuracy is defined by the following.
	\begin{enumerate}
	  	\item \textbf{Time in Goal}, how long the plant was able to hold the endpoint within the target location.
	\end{enumerate}
\end{enumerate}

\subsection{Loss functions}

To induce specialisation, we used two loss functions: One to represent specialisation seen in the dominant hemisphere, `coordinated, efficient movement', and one for the non-dominant hemisphere `positionally accurate and stable'. To achieve this, four loss functions were used in combination, described in the following paragraphs and shown with their relative weights in Table~\ref{tab:loss}. The weights were hyperparameters determined and optimised empirically except for the baseline (non-specialised) models and the combining/output layers of the specialised models, where the Dominant and Non-dominant loss functions were combined in a 50:50 split.

\paragraph{Cartesian 1 (absolute position loss) and Cartesian 2 (squared position loss)}
Previous studies have shown that in motor learning, the loss function for small errors in the dominant hand approximates the squared position error \cite{kording_loss_2004}, whereas for larger errors it is closer to linear, making it more resilient to outliers. The use of these functions (Cartesian 2 for the dominant limb, and Cartesian 1 for the non-dominant limb) corresponds to the greater accuracy in movement seen for the dominant hand and the greater stability seen in the non-dominant hand.

\paragraph{Muscle Activation Loss}
Used in the DL system to increase coordination and efficiency of movement, similar to what is observed in the dominant system. Some modelling approaches have used this assumption in optimisation (minimisation of summed muscle forces) \cite{erdemir_model-based_2007}.

\paragraph{Weight Penalty}
An additional weight penalty was added to the NDL system to build a simpler approach as tasks favouring the non-dominant limb are often conducted on `autopilot' rather than requiring active thought (such as holding something steady while performing a more complex task with the dominant hand).

\begin{table}
     \centering
     \newcolumntype{P}[1]{>{\centering\arraybackslash}p{#1}}
     \begin{tabular}{
     P{0.2\columnwidth}
     P{0.14\columnwidth}
     P{0.14\columnwidth}
     P{0.15\columnwidth}
     P{0.14\columnwidth}}
          \toprule
          Loss Hemisphere & Cartesian 1 (absolute)  & Cartesian 2 (squared) & Muscle Activation & Weight Penalty \\
          \midrule
          DL & 0 & 5 & 2 & 0 \\
          NDL & 5 & 0 & 0 & 2 \\
          Combined & 2.5  & 2.5 & 1 & 1 \\
  		  \bottomrule
     \end{tabular}
     \caption{Loss Function Weights. DL and NDL are the dominant and non-dominant loss functions, respectively. Cartesian 1 refers to absolute position loss, Cartesian 2 to squared position loss.}
     \label{tab:loss}
 \end{table}

\subsection{Network architecture}

We created three network structures: Unilateral, Bilateral, Bilateral with Corpus Callosum; an example is shown in Figure~\ref{fig:arm_arch}b. Two main hyperparameters describe the structure of all the networks: 1) Units (U), the number of neurons per layer and 2) Layers (L), the number of hidden layers before.
All networks have an output layer that is (M) neurons wide, where M is the number of muscles (6). 
It uses the sigmoid activation function to scale the result between 0 (no muscle activation) and 1 (full force muscle activation). 
The hidden layers use a tanh activation function, which was found to give the best performance. 

\subsubsection{Unilateral Network}
A simple multilayer perceptron (MLP) used as a baseline and to tune the loss functions. All hidden layers are densely connected.  
 The unilateral model can be trained without specialisation (Uni-B) or with a `dominant system' loss function (Uni-DL) or non-dominant (Uni-NDL).

\subsubsection{Bilateral Network}
The Bilateral Network has two hemispheres without inter-hemispheric communication, and the outputs are combined to produce the model output. This network can be trained specialised (Bi-S) or non-specialised (Bi-NS) and can be used for exploring specialisation or as a baseline. To ensure an equivalent number of neurones across models, each hidden layer contains (U/2) neurons per side. Each side contains (L) hidden layers.
 
The Combination Layer is a weighted addition of the L-th hidden layer from each hemisphere. Because of this, it also has (U/2) neurons. For this combination layer, the two weights are trainable parameters. This structure has less trainable weights than the baseline model, with the trade-off of having the same neuron count. This means the models are not perfectly comparable; however, the number of trainable parameters is close enough to compare.

\subsubsection{Bilateral Network with Corpus Callosum}
This model adds inter-hemispheric communication. Like the Bilateral Network this can be trained in both a specialised and non-specialised way and so is used for exploring specialisation and baseline performance.
 
Communication between the hemispheres is achieved through pooled representations of the layers from the opposing hemisphere. The pooled representation is 50\% (rounded down) of the size of the layer it represents and is added directly to the output of the corresponding layer from the other hemisphere before dense connections to the next layer. 50\% is much higher than what is observed in the human brain \cite{drake2009gray}, however a lower percentage than this would have resulted in only single neurons being affected across our relatively light models, so we accepted the tradeoff. As the communication is achieved through pooling layers, there are no additional trainable parameters over the Bilateral Network. Combination and output layers are the same as in the Bilateral Network. The model can have specialised (CC-S) or non-specialised (CC-NS) hemispheres.

\subsection{Training process}
For the Unilateral (Uni-B, Uni-DL, Uni-NDL) and Non-Specialised (Bi-NS, CC-NS) models, the training followed a supervised learning process with targets defined by the task. We used a maximum epoch count of 100, which was rarely reached due to the early stopping criterion which monitored the validation loss value and terminated training after seeing no improvement over 3 epochs. The training was carried out on 10 random seeds.

The specialised models (Bi-S, CC-S) required a different approach to induce specialisation via different loss functions. In this case, for each training step, the gradients for the DL and NDL losses were calculated separately and then combined in a weighted addition before being used to update the trainable parameters. The weighted addition of the gradients ensures that gradient matrices relating to the layers of the Dominant Hemisphere were comprised solely from the DL gradient, gradient matrices relating to the layers of the Non-dominant Hemisphere were comprised solely from the NDL gradient and the gradient matrices relating to the remaining `combination/output layers' were a 50:50 split of both the DL and NDL gradients. Otherwise, training was identical to the non-specialised models including maximum epoch count and early stopping rules.

\subsection{Lesion testing}

\begin{figure}
    \centering
    \includegraphics[width=0.7\linewidth]{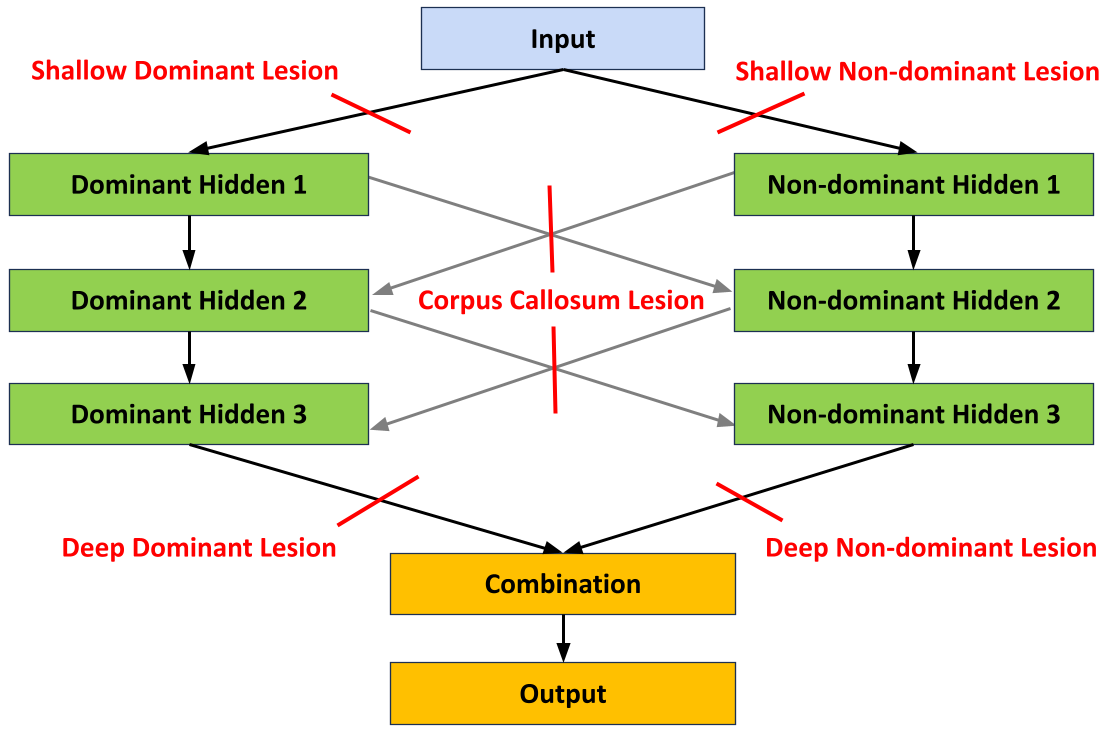}
    \caption{Lesion Positions across a Bilateral Network}
    \label{fig:lesions}
\end{figure}

Lesion studies are commonly used to understand cognitive and computational models \cite{meyes_ablation_2019}, as they allow us to observe how parts of the model behave independently of the whole.
 There are many different places to lesion the networks. We chose 7, each serving different purposes (see Figure~\ref{fig:lesions}). The Shallow lesions cut off the input signal to the hemisphere, however still allow for the weights and biases between the layers to pass down what can be thought of as a `default signal'. The Deep Lesions cut off all the hemisphere's contribution to the output layer; however, in the case of the Corpus Callosum network will still allow healthy signals to flow to the opposing hemisphere. The Corpus Callosum lesion removes the communication between the two hemispheres and enforces the model to behave identically to the Bilateral Network. The final two lesions combine the Corpus Callosum Lesion with the Deep lesions from each hemisphere, ensuring that the healthy hemisphere is completely independent of the lesioned hemisphere.

As lesioning essentially results in half of the model being removed, the resulting outputs are severely hampered to the point where outputs were less meaningful. For this reason, we found that freezing everything except the output layer and allowing the model to train for a single epoch after lesioning gave more insightful results. For this step, all models/lesion types used the same combined loss function, Table~\ref{tab:loss}.

\section{Results}

Before running tests, we empirically optimised hyperparameters: U=10, L=2, activation function (hidden layers) = tanh, learning rate = 0.001, optimiser = Adam, task timesteps = 50 (both tasks).
Larger models were also tested; however, we found that for our Unilateral Network, performance gains were marginal, and for all of the Bilateral Networks (incl. Corpus Callosum) we found that the models were prone to falling into particular local minima where a single hemisphere would become strong enough to dominate the other. With regularisation, the weights of the other hemisphere were forced to zero. This did not allow exploration of bilateralism. We hypothesise that this is due to the simplicity of the tasks and the fact that larger models are overpowered for the task.



\subsection{Specialisation in the Unilateral Network}
We first investigated whether using different loss functions was an appropriate way to achieve the desired specialisation. The tests were conducted with a unilateral architecture.

\begin{figure*}
    \centering
    \begin{subfigure}[]{0.5\textwidth}
        \centering
	    \includegraphics[height=1.3in]{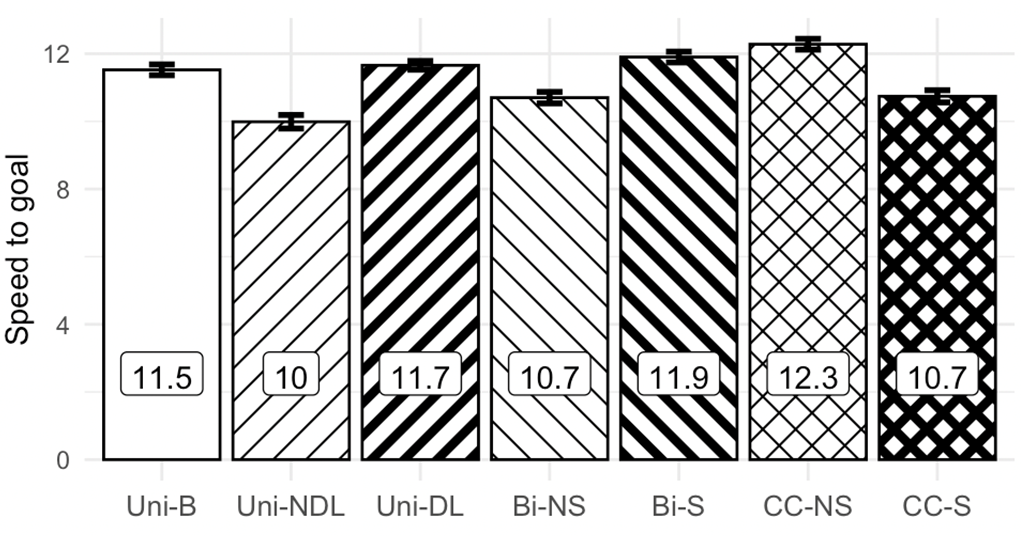}
    	\caption{Random Reach Task (Dominant system is preferred)}
	    \label{fig:all_reach}
    \end{subfigure}%
    ~ 
    \begin{subfigure}[]{0.5\textwidth}
        \centering
	    \includegraphics[height=1.3in]{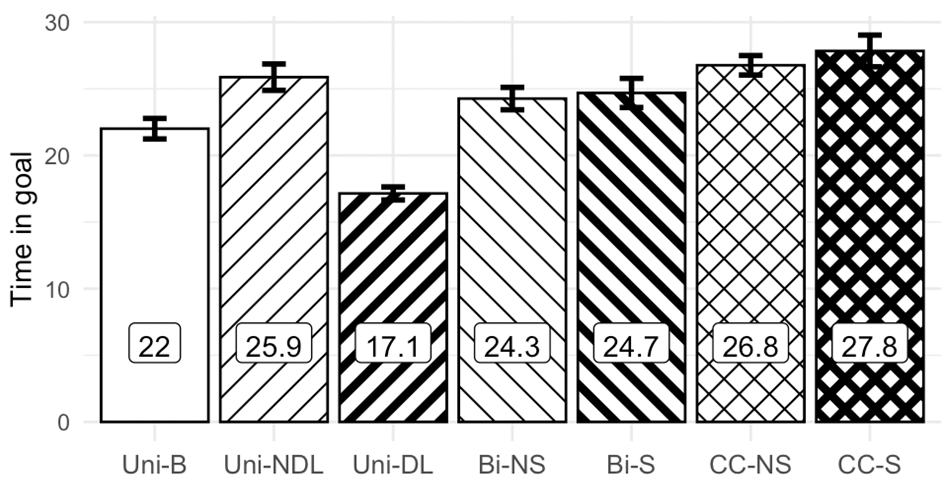}
    	\caption{Hold Position Task (Non-Dominant preferred)}
	    \label{fig:all_hold}
    \end{subfigure}
    \caption{All model performance on both tasks. Each system is trained with a loss function suited to one of the tasks, resulting in a `preferred' system per task.}
    \label{fig:all_models}
\end{figure*}

In the Random Reach Task (chosen to favour the Dominant hemisphere), both loss functions are effective in training a model that can complete the goal (96.7\% Uni-DL, 94.6\% Uni-NDL), however the Uni-DL model on average completes the task faster than the Uni-NDL model resulting in a $\sim$17\% efficiency gain in this metric, Figure~\ref{fig:all_models}. The results of the Hold Position Task (chosen to favour the Non-dominant hemisphere) show a wider margin between the models, with Uni-NDL able to retain its position inside the goal $\sim$50\% longer than Uni-DL.

The results show that we are able to achieve the specialisation that is seen in the hemispheres of the human brain by training the model using loss functions tailored to the desired behaviour.

%
%

\subsection{Specialisation in the Bilateral Network}
\begin{figure*}[t]
    \centering
    \begin{subfigure}[]{1.\textwidth}
    \includegraphics[width=1.0\linewidth]{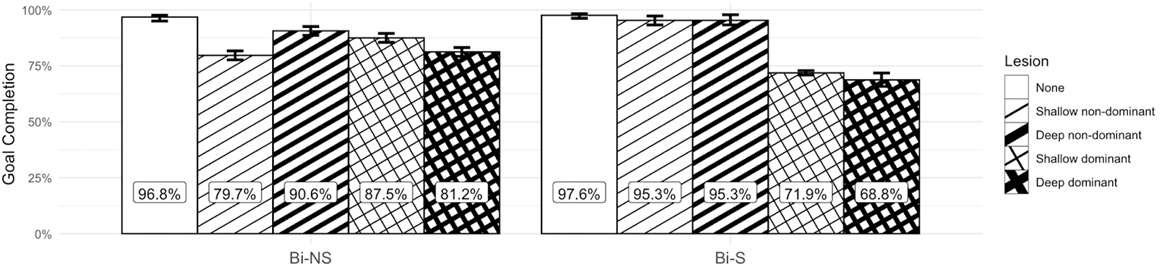}
    \caption{\textbf{Random Reach Task (Dominant System is preferred).} The specialised Dominant hemisphere (Non-dominant lesion) are comparable to the non-lesioned, outperforming the Non-dominant. Difference between hemispheres is minimal in the Non-specialised model.}
    	\label{fig:lesion_reach}
    \end{subfigure}

    \begin{subfigure}[]{1.\textwidth}
    \includegraphics[width=1.0\linewidth]{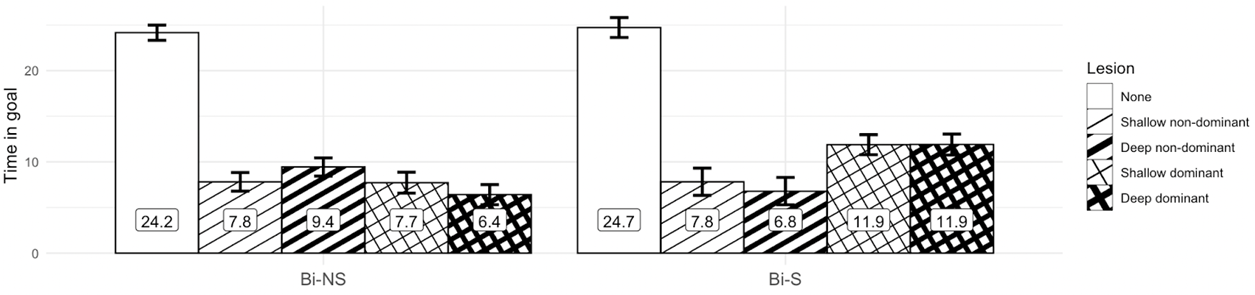}
    \caption{\textbf{Hold Position Task (Non-dominant System is preferred).} The specialised Non-Dominant hemisphere (Dominant lesions) outperforms the Dominant hemisphere. Difference between hemispheres is minimal in the Non-specialised model.}
    	\label{fig:lesion_hold}
    \end{subfigure}
    \caption{Lesion performance of Bi-S and Bi-NS}
    \label{fig:lesions_bi}
\end{figure*}

Given the success of the loss functions, a bilateral model was trained using the two loss functions to train each hemisphere of the model.
 To test the level of specialisation across the hemispheres, the trained model was lesioned in different places and the performance of the model was observed and compared to a non-specialised bilateral model as a baseline, Figure~\ref{fig:lesions_bi}.
Results labelled as `Non-dominant Lesion' show performance of the Dominant Hemisphere and results labelled `Dominant Lesion' show the performance of the Non-dominant Hemisphere.

%

As expected, the Bi-NS model shows little hemispheric specialisation toward either task. For the Bi-S model, on both tasks, the preferred hemisphere outperforms both hemispheres in the Bi-NS model, which outperforms the non-preferred hemisphere from the Bi-S model.

\subsection{Specialisation in the Bilateral Network with Corpus Callosum}
When comparing shallow lesions with deep lesions in models employing a Corpus Callosum, we can see that in both the CC-S and CC-NS models, shallower lesions impact the ability of the model much more than deeper lesions. We theorise that this is due to the fact that when performing a deep lesion, healthy neuron signals are still flowing from the damaged hemisphere to the healthy hemisphere providing the healthy hemisphere with usable data. However, when shallow lesions exist, damaged signals are also carried through the Corpus Callosum compromising the healthy hemisphere's ability.

\begin{figure*}[h!]
    \centering
    \begin{subfigure}[]{1.\textwidth}
    \includegraphics[width=1.0\linewidth]{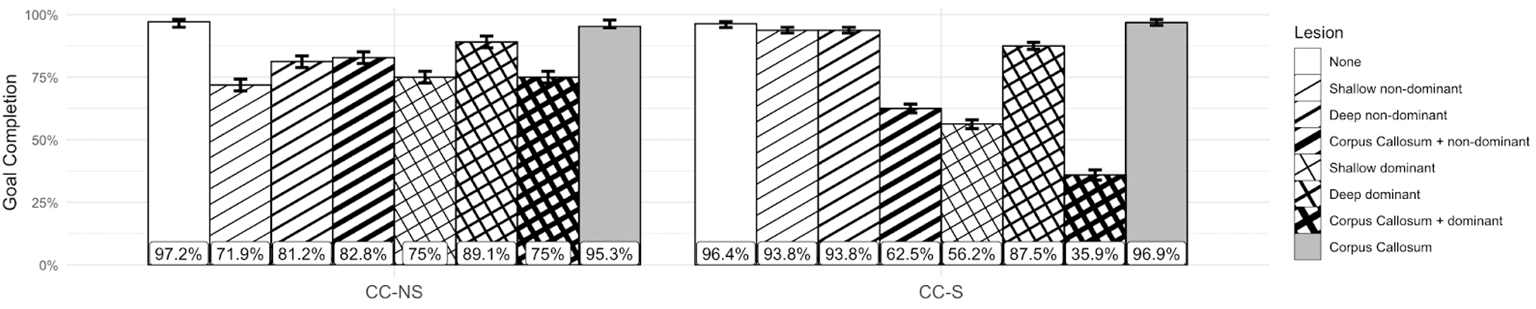}
    \caption{\textbf{Random Reach Task (Dominant System is preferred).} Model shown to regain most of its performance after Corpus Callosum lesion. Specialisation see to be achieved, with the Dominant hemisphere performance (Non-dominant lesions) performing better than the Non-dominant hemisphere.}
    \label{fig:cc_lesion_reach}
	\end{subfigure}
	
	\begin{subfigure}[]{1.\textwidth}
    \centering
    \includegraphics[width=1.0\linewidth]{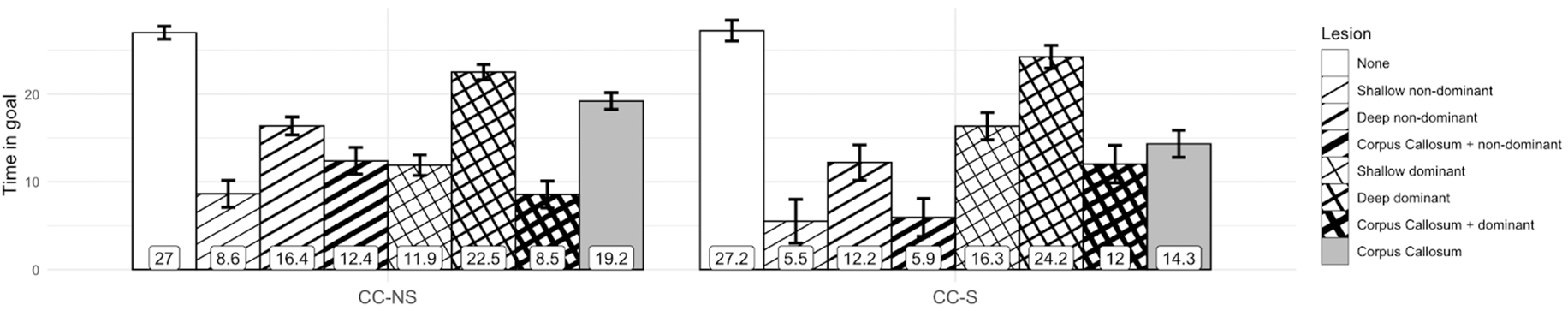}
    \caption{\textbf{Hold Position Task (Non-dominant System is preferred).} Corpus Callosum lesion shown to severely impact the performance of the model. Deep lesions are less impactful than the Shallow lesions due to information flow still present through the Corpus Callosum.}
    \label{fig:cc_lesion_hold}
	\end{subfigure}
	\caption{Lesion performance of CC-S and CC-NS.}
	\label{fig:lesions_cc}
\end{figure*}

The tests show an apparent specialisation in the network hemispheres with each system performing well for the preferred task, i.e. dominant outperforming non-dominant in the Reach Task, Figure~\ref{fig:cc_lesion_reach}, and non-dominant outperforming dominant in the Hold Position Task, Figure~\ref{fig:cc_lesion_hold}. In the Reach task, this specialisation is less apparent when looking at the deep lesioned models, however, could be due to the reasons given in the previous paragraph.

Removal of the Corpus Callosum showed a different effect in each task. For the Reach task, after the single re-training epoch, the model was able to regain very close to all of its performance on the task when compared to the non-lesioned model. However, in the Hold Position Task, removal of the Corpus Callosum resulted in a large decrease in performance, much more comparable to that seen when performing a shallow dominant side lesion. This suggests that the non-dominant hemisphere, while specialised to perform better at the Hold Position Task, is still leveraging signals from the dominant hemisphere to aid in decision making. On the other hand, the ability of the model to regain the majority of its performance on the Random Reach task after retraining suggests that the dominant hemisphere alone is sufficient to complete the task at a satisfactory level, making the non-dominant hemisphere more redundant to the task.

\subsection{Overall performance}

For comparison of model performance, refer to Figure~\ref{fig:all_models} and Post Hoc ANOVA analysis, Appendix, Figure~\ref{fig:anova}.
On the Random Reach Task we focussed on Speed to Goal, as all models scored above 95\% for Goal Completion (Appendix, Figure~\ref{fig:rrtask_gc}). The specialised bilateral models did not conclusively outperform the non-specialised. In terms of speed to goal, the CC-NS model was the strongest; however, when looking at goal completion the Bi-S model was best. In the Hold Position Task, the specialised models outperform their non-specialised equivalents; however, they are not statistically significant. CC-S was significantly better than any other model.
All bilateral models (Bi-NS, Bi-S, CC-NS CC-S) were statistically significantly better than the unilateral baseline (Uni-B).
In both tasks, the best specialised bilateral model had either slightly better performance to the preferred side (Bi-S vs Uni-DL for Random Reach Task) or significantly better (CC-S vs Uni-NDL for Hold Task); and significantly better than the non-preferred one.
In both tasks, the best model had a Corpus Callosum, indicating its benefit. 

\section{Discussion}
We found that using differential loss functions, it is possible to effectively induce specialisation.
Furthermore, it was clear that specialisation improves performance on relevant tasks.
This is in accordance with `Complementary Dominance' \cite{woytowicz_handedness_2018} where each hemisphere is specialised for different aspects of motor control, as opposed to the `dominant side' being superior in all tasks (Global Dominance).

The results suggest there are two ways that specialisation could be exploited in a bilateral architecture.
First, the hemispheres complement each other to improve performance on a given task, as shown by CC-S in the Hold Position Task.
This is in agreement with work where in the event of perturbations in either arm (dominant/non-dominant), an early EMG response was observed in the non-dominant anterior deltoid suggesting that the non-dominant system aids in the response to perturbations in the dominant system \cite{schaffer_interlimb_2021}.
Second, a bilateral architecture is capable of different tasks, and the preferred hemisphere could be selected in a task-dependent manner, as in Sainburg's \cite{sainburg_handedness_2005} theory on handedness. This was shown by the fact that the best specialised bilateral models had approximately equal performance to the unilateral model specialised for the task (despite half the number of trainable parameters). As a bilateral model has differentially specialised hemispheres, they are suited to operating on both task types, whereas a unilateral model is limited to one.

\section{Limitations and future work}
The architecture takes inspiration from biology, but the models do not mimic important neurobiological characteristics such as substrate differences between the hemispheres, as modelled in the studies mentioned in the background and reviewed in \cite{rajagopalan2022deep}. Including those features and comparing their performance with human studies is an interesting direction for future work.
In addition, the models were trained on quite simplistic tasks, and this may not elucidate the models' full capabilities. 
Further research could introduce more complex tasks. 
Increasing model and task complexity could include a system that controls two arms, each primarily controlled by one hemisphere. 
%
%
The robustness of the architectures may also be explored through the addition of noise to the model input. 

\section{Conclusion}


We were successfully able to induce specialisation into our models through training each hemisphere on different loss functions, tailored toward the desired behaviour, emulating what is observed in humans, where the dominant side of the body is more coordinated and efficient with its movements and the non-dominant side shows better positional accuracy and resilience to external perturbations. 
We tested our models on a Random Reach Task and a Hold Position Task, suited to dominant and non-dominant systems respectively. 
Specialisation conferred benefits on specific tasks and could be exploited in a bilateral architecture.
The results provide a foundation for building bilateral models to better understand the brain's relation to motor control and inspire novel approaches to AI motor control.


\bibliographystyle{splncs04}
\bibliography{references}

\appendix
\section{Appendix}

\begin{figure}
    \centering
    \includegraphics[width=0.5\linewidth]{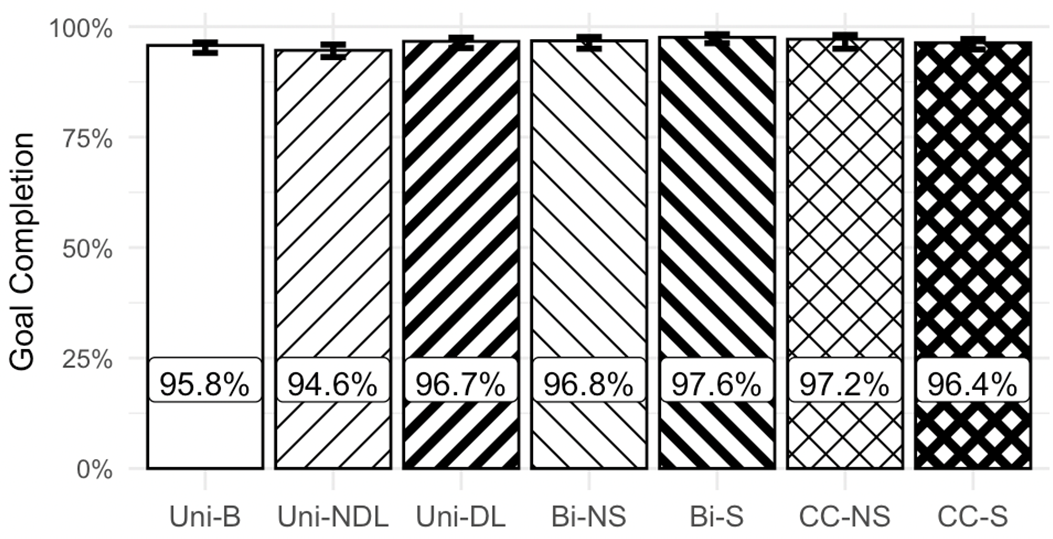}
    \caption{Random Reach Task - Goal Completion (Dominant System is preferred)}
    \label{fig:rrtask_gc}
\end{figure}

\begin{figure}
\begin{subfigure}{0.7\textwidth}
    \centering
    \includegraphics[width=\textwidth]{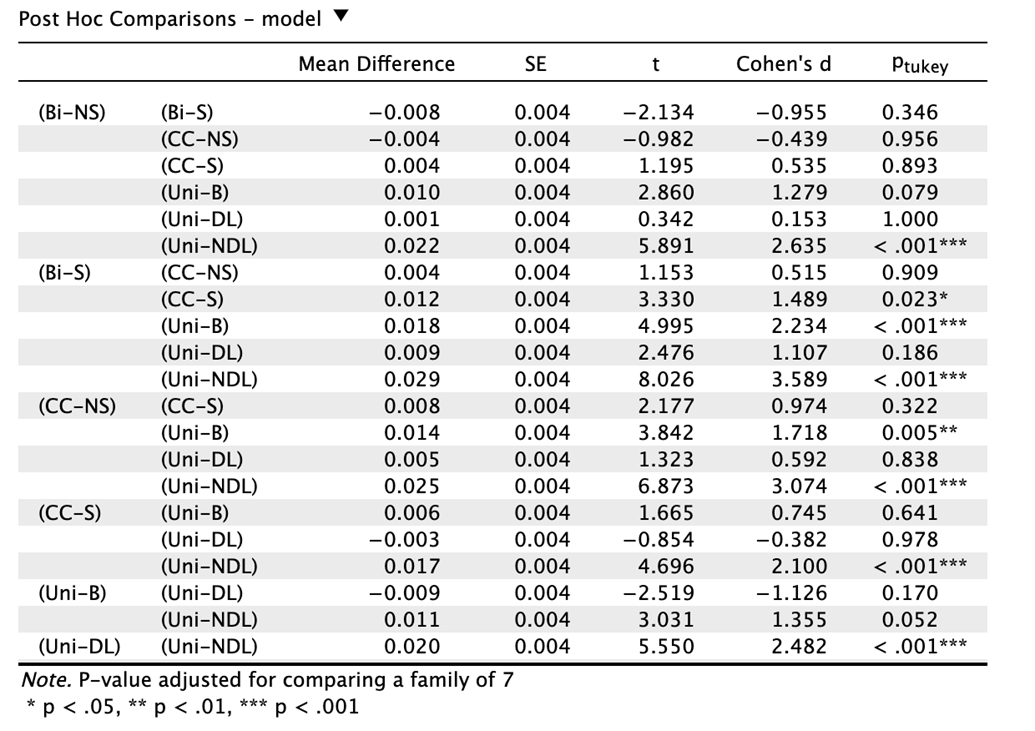}
    \caption{Random Reach Task -- Goal Completion}
    \label{fig:anova_goal_completion}
\end{subfigure}
~
\begin{subfigure}{0.7\textwidth}
    \centering
    \includegraphics[width=\textwidth]{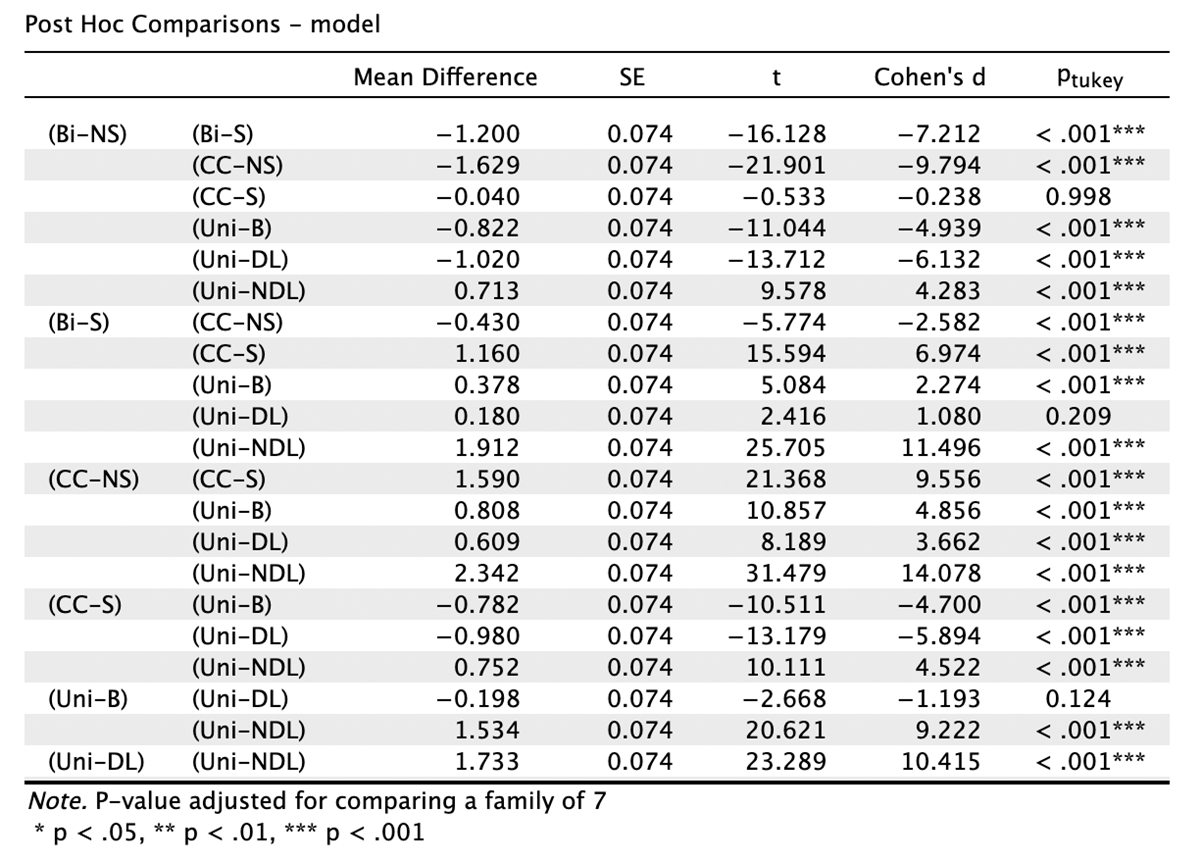}
    \caption{Random Reach Task -- Speed to Goal}
    \label{fig:anova_speed}
\end{subfigure}
~
\begin{subfigure}{0.7\textwidth}
    \centering
    \includegraphics[width=\textwidth]{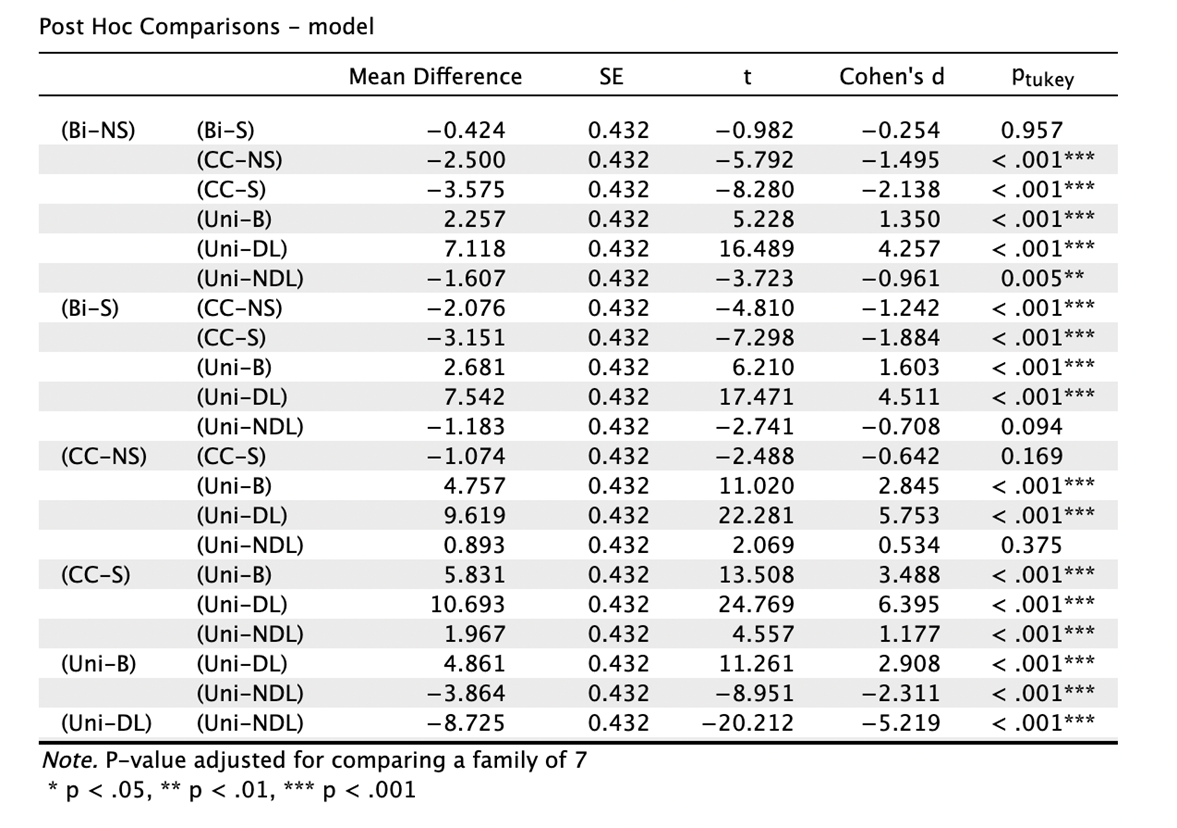}
    \caption{Hold Position task -- Time in Goal}
    \label{fig:anova_time}
\end{subfigure}
\caption{ANOVA post-hoc pairwise analysis.}
\label{fig:anova}
\end{figure}


\end{document}